\documentclass[sigconf,dvipsnames]{acmart}
\usepackage{booktabs}
\usepackage{listings}
\usepackage{lipsum}

\graphicspath{{figs/}}

\AtBeginDocument{%
  \providecommand\BibTeX{{%
    \normalfont B\kern-0.5em{\scshape i\kern-0.25em b}\kern-0.8em\TeX}}}

\setcopyright{licensedusgovmixed}
\copyrightyear{2021}
\acmYear{2021}
\acmISBN{978-1-4503-8691-3/21/07}
\acmDOI{10.1145/3477145.3477155}

\acmConference[ICONS '21]{ICONS 2021: International Conference on Neuromorphic Systems}{July 27--29, 2021}{PREPRINT}
\acmBooktitle{PREPRINT}

\begin{document}

\title{Neko: a Library for Exploring Neuromorphic Learning Rules}


\author{Zixuan Zhao}
\affiliation{%
  \institution{University of Chicago}
  \country{}
}

\author{Nathan Wycoff}
\affiliation{%
  \institution{Virginia Tech}
  \country{}
}

\author{Neil Getty}
\affiliation{%
  \institution{Argonne National Laboratory}
  \country{}
}

\author{Rick Stevens}
\affiliation{%
 \institution{Argonne National Laboratory \& University of Chicago}
  \country{}
}

\author{Fangfang Xia}
\affiliation{%
 \institution{Argonne National Laboratory \& University of Chicago}
  \country{}
}


\begin{abstract}
The field of neuromorphic computing is in a period of active exploration.
While many tools have been developed to simulate neuronal dynamics or convert deep networks to spiking models, general software libraries for learning rules remain underexplored.
This is partly due to the diverse, challenging nature of efforts to design new learning rules, which range from encoding methods to gradient approximations, from population approaches that mimic the Bayesian brain to constrained learning algorithms deployed on memristor crossbars.
To address this gap, we present Neko, a modular, extensible library with a focus on aiding the design of new learning algorithms.
We demonstrate the utility of Neko in three exemplar cases: online local learning, probabilistic learning, and analog on-device learning.
Our results show that Neko can replicate the state-of-the-art algorithms and, in one case, lead to significant outperformance in accuracy and speed.
Further, it offers tools including gradient comparison that can help develop new algorithmic variants.
Neko is an open source Python library that supports PyTorch and TensorFlow backends.
\end{abstract}



\begin{teaserfigure}
  \includegraphics[width=\textwidth]{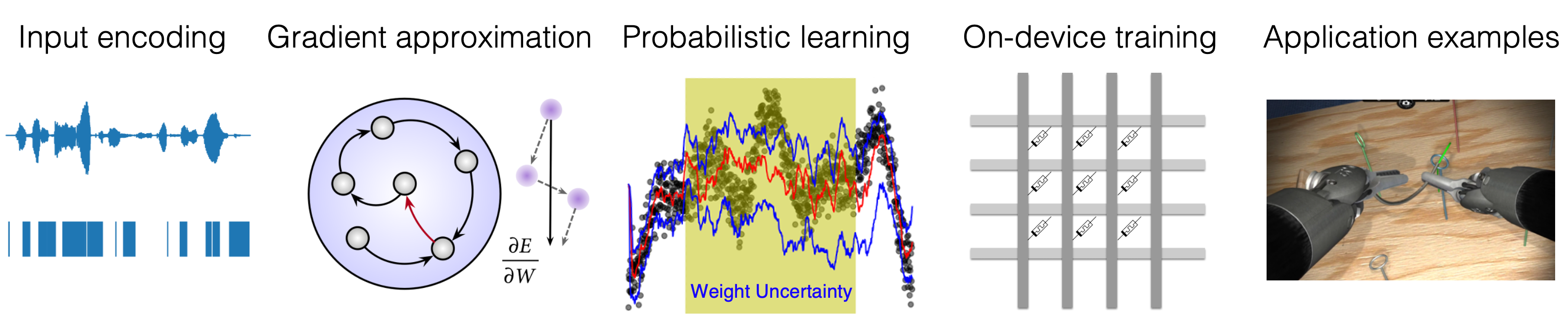}
  \caption{Neko overview. \textmd{Key components in the neuromorphic learning library.}}
  \Description{Key components in the Neko neuromorphic learning library}
  \label{fig:teaser}
\end{teaserfigure}

\maketitle


\section{Introduction}
Deep learning is the prevailing paradigm for machine learning.
Over the course of its meteoric rise, its many differences from human learning have become increasingly clear.
Chief among these are gaps in data efficiency, robustness, generalizability, and energy efficiency --- all unlikely to narrow with growing computation power alone.
This has motivated a renewed search for brain-inspired learning algorithms.
However, the current software infrastructure needs improvement to support productive exploration.

Two common choices today for designing novel learning algorithms are TensorFlow \cite{abadi2016tensorflow} and PyTorch \cite{paszke2019pytorch}.
These general deep learning frameworks provide powerful abstractions for calculating gradients and building deep neural networks, but there is no intermediate layer between these two levels.
For high-level development, backpropagation is the only learning algorithm offered and is in fact coupled with the training process.

Software in neuromorphic computing, on the other hand, has traditionally focused more on simulating neurons and spiking neural networks \cite{carnevale2006neuron,gewaltig2007nest,bekolay2014nengo,stimberg2019brian}, interfacing with neuromorphic hardware \cite{davison2009pynn,sawada2016truenorth,lin2018programming,rueckauer2021nxtf}, and converting pre-trained deep learning models to spiking neural networks for inference \cite{rueckauer2017conversion,rueckauer2018conversion}.
Learning has not been a key part of these libraries.
The few supported learning rules such as spike-timing-dependent plasticity are not competitive on large problems.
As a result, new learning algorithms are developed in independent codebases that are not easily reusable.

In this work, we present Neko, a software library under active development for exploring learning rules.
We build on the popular autograd frameworks, and our goal is to implement key building blocks to boost researcher productivity.
By decoupling the learning rules from the training process, we aim to provide an abstraction model that enables mixing and matching of various design ideas.
To arrive at the right abstraction level, we need to sample a wide range of learning algorithm research.
Below are the three directions and exemplars we have prioritized in this initial code release.

The first class of learning rules are gradient-based methods.
They approximate backpropagation with various levels of biological plausibility \cite{lillicrap2020backpropagation,lee2016training,sacramento2018dendritic,neftci2019surrogate,zenke2018superspike,marschall2020unified,lillicrap2016random,akrout2019deep,sornborger2019pulse}.
From this category, we study the e-prop algorithm \cite{bellec2020solution} in detail and provide a complete reimplementation.
The second direction is based on the hypothesis that the brain keeps track of probabilistic distributions over weights and rewards \cite{aitchison2021synaptic,dabney2020distributional}.
This line of exploration may offer important clues towards achieving learning efficiency and robustness in the face of uncertainty.
We develop a sampling-based learning rule on spiking neural networks (SNN).
The third class is concerned with hardware constraints on plasticity mechanisms.
For this class, we include the classic example of Manhattan rule training for memristive crossbar circuits.
In all three exemplars, we seek consistent implementation in the Neko library.

\section{Library design}
The Neko library is designed to be modular, extensible, and easy to use.
Users can select from a collection of neuron models and encoding methods to build a spiking or regular artificial neural network, and train it with one of the implemented learning rules.
Alternatively, they could supply their own networks from PyTorch or Keras \cite{chollet2015keras} or develop new learning algorithms based on the provided intrinsics.
The following code snippet provides an example of solving MNIST \cite{lecun1998mnist} with the e-prop algorithm on a recurrent network of 128 hidden adaptive leaky integrate-and-fire (ALIF) neurons.

\begin{lstlisting}[caption={Train an SNN model of ALIF neurons with e-prop. },captionpos=b,frame=single, language=python,breaklines]
from neko.backend import pytorch_backend as backend

rsnn = ALIF(128, 10, backend, task_type='classification')
model = Evaluator(rsnn, loss='categorical_crossentropy', metrics=['accuracy', 'firing_rate'])
learning_rule = Eprop(model, mode='symmetric')
trainer = Trainer(learning_rule)
trainer.train(x_train, y_train, epochs=30)
\end{lstlisting}

The training process illustrated in this example can be broken down into a series of high-level Neko modules:
the \emph{layer} includes pre-implemented recurrent SNNs and adaptors for existing Keras and PyTorch models;
the \emph{evaluator} associates a model with a loss function and optional metrics;
the \emph{learning rule} implements backpropagation and a growing list of neuromorphic learning rules;
and the \emph{trainer} handles training logistics as well as special logic to apply multiple learning rules for gradient comparison between models.
Besides these core components, auxiliary modules include the data loader, spike encoder, optimizer, and functions for loss, activation, and pseudo-derivatives calculations.

To help users define custom algorithms, Neko also provides a unified API for accessing frequently used features in TensorFlow and PyTorch such as low-level tensor operations.
Switching the backend is straightforward.
This feature can detect occasional framework-dependent behavior and is useful for code verification and performance analysis.
The multi-backend support is reminiscent of the earlier Keras framework.
However, Neko is different in that it provides more fine-grained abstraction layers such that users can replace the learning algorithm by changing a single line of code.
Taken together, these features also simplify the process of porting code to hardware accelerators, since implementing a backend for the hardware is sufficient to run all models in Neko on it.

\section{Use cases}
In this section, we present results on the three representative learning rules introduced earlier.
We also provide gradient analysis as an example of Neko's cross-cutting utilities that we are building to help design, debug, and compare new learning algorithms.

\subsection{Credit assignment with local signals}
A key mystery in the brain is how it implements credit assignment.
The standard backpropagation through time (BPTT) algorithm is unrealistic as we cannot expect a biological neuron to be aware of all past synaptic strengths.
Bellec et al. \cite{bellec2020solution} proposed e-prop, a local online learning algorithm for recurrent SNNs.
The method exploits the mathematical formula of BPTT, deriving an approximation which only requires a recursive accumulative \emph{eligibility trace} and a local \emph{learning signal}.
These properties make the algorithm one step closer to biologically realistic on-chip learning.

In Neko, we implemented full-featured e-prop algorithms including the three variants: symmetric, random, and adaptive.
Whereas the paper manually derived the e-prop formulas for some networks, we took a different approach: separating the model from the learning rules.
In the layer module, the regular recurrent neural networks and recurrent SNNs, with leaky integrate-and-fire (LIF) or ALIF neurons, were all defined as standard models.
Meanwhile, they inherited from an \emph{Epropable} class, which defined general symbolic gradient formulas according to recurrent cell dynamics.
Specifying this extra information was all it took to perform e-prop, and in a network-agnostic way.
This design enabled the error-prone formula derivation to be automated.
It also sped up experiments with new network architectures or e-prop variants.

\begin{figure}
    \centering
    \includegraphics[width=\linewidth]{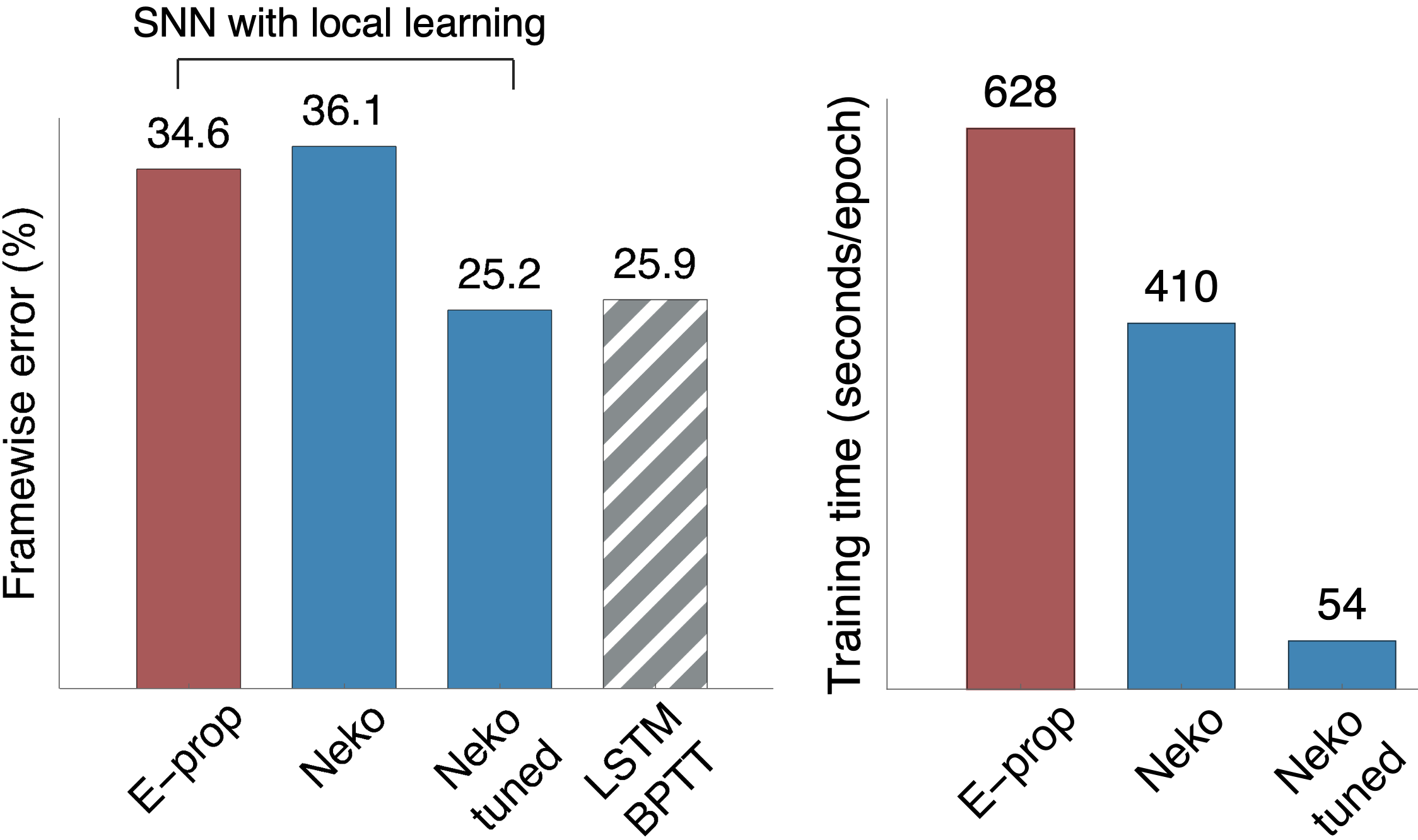}
    \caption{TIMIT results.
      \textmd{We reproduce e-prop accuracy on speech recognition in Neko with a smaller network. Neko is faster with slight tuning and reduces error by $27\%$ to reach the nonspiking baseline performance of a BPTT-trained LSTM model.}
    }
    \label{fig:timit}
\end{figure}

We compared the Neko implementation of e-prop to the original implementation on the TIMIT benchmark \cite{garofolo1992timit} for framewise speech recognition.
The authors reported the results on a hybrid network of 100 ALIF and 300 LIF neurons \cite{bellec2020solution}.
In our experiment, we used an ALIF-only network of 200 neurons and otherwise kept the setup identical.
We report close reproduction accuracy in Fig. \ref{fig:timit}.
Notably, Neko's error rate dropped by $27\%$, after tuning regularization and batch size, while keeping the firing rate low at 10 Hz.
To the best of our knowledge, this is the best SNN accuracy obtained with a local learning rule, which in fact reaches the level of an LSTM baseline trained with the precise gradients from BPTT (\cite{bellec2020solution} Fig. S4).
Additionally, Neko is faster (training time from Nvidia V100) and convenient for iterative development.

\subsection{Probabilistic learning}
Bayesian statistics has captured much attention in the computational neuroscience community, both as an explanation for neural behavior \cite{Knill2004} as well as a means of performing inference in neural networks. In Neko, we develop a Hybrid Monte Carlo, or HMC \citep{Neil2011HMC}, algorithm to perform Bayesian inference on spiking neural networks based on Metropolis-adjusted Langevin diffusion \cite{Rossky1978}.

Fundamentally, HMC algorithms are simply Metropolis-Hastings samplers \cite{Hoff2009Bayes} where the proposal distribution is based on the gradient. Though spiking neurons are non-differentiable by definition, \textit{surrogate gradients} can be defined by considering smoothed versions of the spiking activation function \cite{neftci2019surrogate}. State of the art learning algorithms for spiking neurons have used these surrogate gradients successfully, and we also find success in deploying them in HMC to form our proposal. In fact, this two-stage approach is especially appealing for spiking neurons, since the theoretical underpinnings of HMC place only very weak restrictions on what the proposal direction should be, and certainly do not require an exact gradient to be satisfied. Thus, from a theoretical perspective, running our algorithm for sufficiently long will result in a sample from our true posterior. Empirically, of course, it is not practical to explore the entire nonconvex, high-dimensional posterior. We therefore verify our implementation numerically.

\begin{figure}
    \centering
    \includegraphics[scale=0.193]{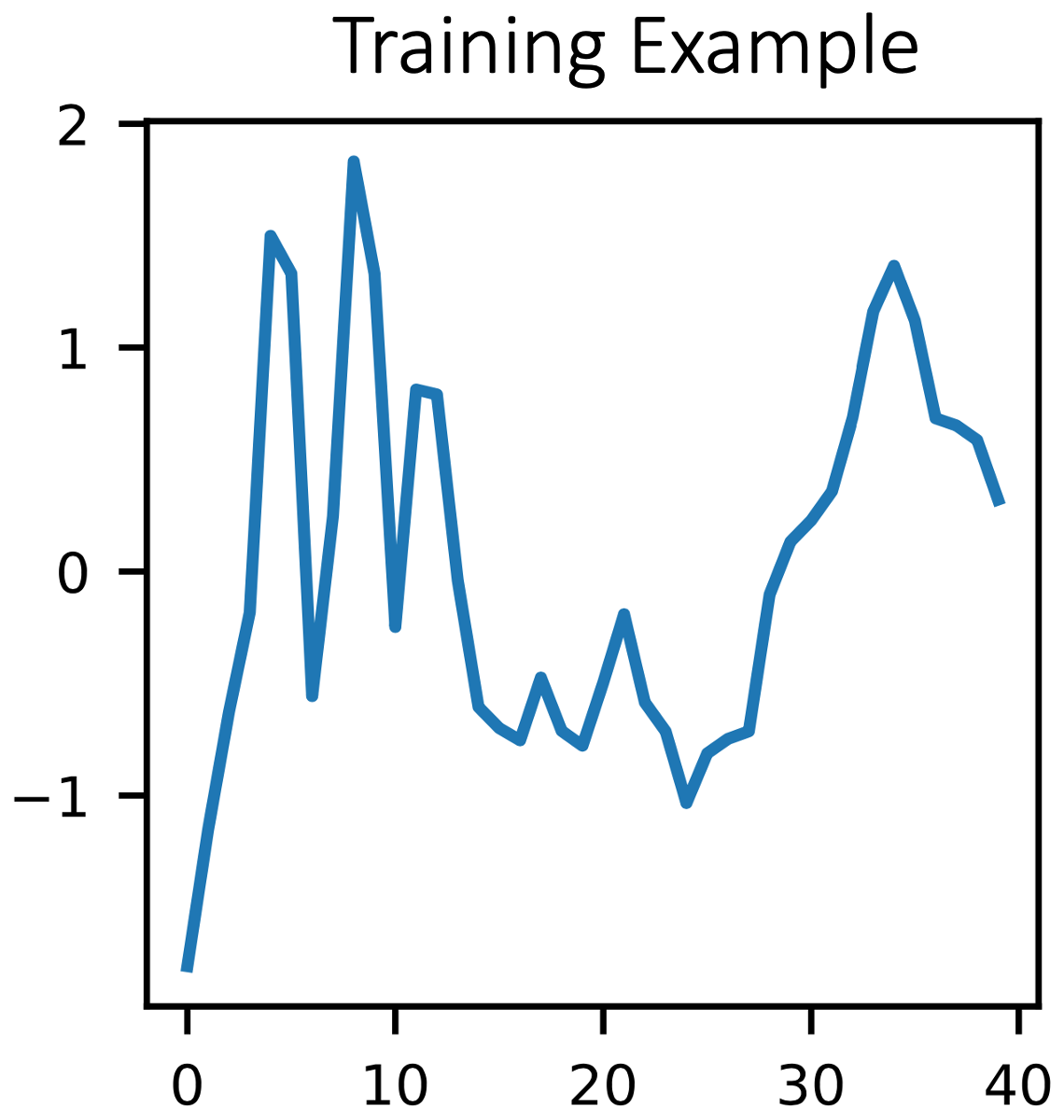}
    \includegraphics[scale=0.8]{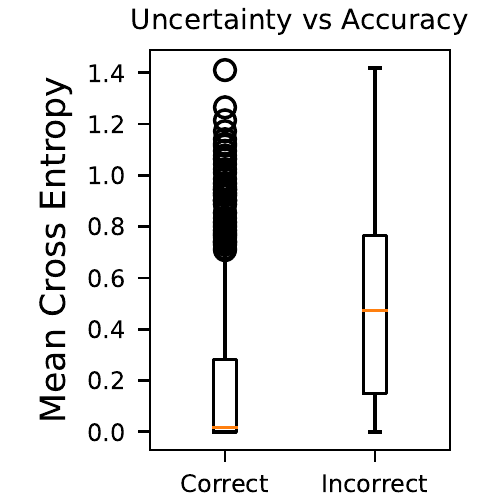}
    \caption{ Uncertainty Quantification.
    \textmd{\textbf{Left:} An example input representing the number 3 for the MNIST-1D data. \textbf{Right:} Posterior uncertainty among test examples which were correctly versus incorrectly predicted. Uncertainty is higher when errors are made.}}
    \label{fig:hmc}
\end{figure}

The MNIST-1D \cite{greydanus2020scaling} data is a derivative of the popular MNIST dataset of handwritten digits which transforms the image recognition problem into a sequence learning problem (See Figure \ref{fig:hmc}, Left). We train a spiking neural network with 1,000 hidden neurons using our proposed HMC algorithm\footnote{Using an adaptive step size \cite{Andrieu2008Adaptive} with a diffusion standard deviation of 0.01 scaled by the norm of the surrogate gradient, which was obtained via standard backpropagation.}, and recorded the posterior mean as well as uncertainty for the train set examples. As shown in Figure 3 (Right), we find that the model displayed significantly more uncertainty on test examples for which its best guess was incorrect than when it was correct. This validates our algorithm, as we would like errors to be associated with high uncertainty.

As future work, we intend to compare HMC and other MCMC algorithms to other probabilistic learning approaches such as Variational Bayes \cite{Graves2011} and Monte Carlo Dropout \cite{gal2016} within the Neko framework.

\subsection{Analog neural network training}
Memristors have emerged as a new platform for neuromorphic learning \cite{thomas2013memristor,hu2014memristor}.
These devices represent the synapse weights in the tunable conductance states of large crossbar architectures.
Compared with digital implementations of neural networks, these analog circuits offer promising advantages in parallel processing, in-situ learning, and energy efficiency \cite{fuller2019parallel,li2018efficient}.
However, they also place constraints on how the weights can be updated.

A classic way to train these networks is with the Manhattan rule learning algorithm \cite{7139171}.
Although training with backpropagation on device is theoretically possible, the time consumption of tuning individual weights with feedback algorithm can be prohibitive, especially for larger scale neural networks \cite{Alibart_2012}.
As an alternative, the Manhattan rule simply updates network weights by a fixed amount according to the sign of the gradients, where the actual change magnitude may depend on the state of the material.
This learning rule has been applied successfully to simple machine learning  benchmarks in simulated or fully hardware-implemented analog neural networks \cite{yao2020fully}.

Neko implements a family of Manhattan rules to simulate the training process.
It includes the basic algorithm and an extended version that supports a specified range of material conductance constraints.
Because these learning rules do not have special requirements for the network architecture, users can directly supply existing Keras and PyTorch models with Neko's adaptors.
Our preliminary results show that both the simple Manhattan rule and the constrained version could train the MNIST dataset up to 96\% accuracy on a simple 2-layer (with 64, 32 neurons) multi-layer perceptron, which is 2\% lower than backpropagation.

\subsection{Gradient comparison analysis}
Many learning rules depend on gradients explicitly or implicitly.
Yet, gradient estimates are not intuitive to developers.
Debugging learning rules sometimes require noticing the subtle differences in gradient estimates and follow their trends over the course of training.
In Neko, we have designed a gradient comparison tool that can enumerate the gradients or weight changes for multiple learning rules with the same model state and input data.
It can also track this information batch by batch.
Visualizing this information can help inspect approximation quality differences caused by algorithm tweaks and identify equivalence in formula transformations.
Outside the context of debugging, the change in gradient estimates throughout the training process can also reveal potential biases and other properties of the learning algorithm.

The gradient comparison tool is made possible by Neko's separation of the learning algorithm and trainer module. It is implemented as a special trainer that takes multiple learning rules and clones of the same model. While the primary model follows the usual training process, the others' parameters are synced with the primary at each training step, and the weight changes are saved. The equivalence of gradient changes and weight changes can be established using the built-in \emph{naive optimizer} which applies gradients directly without learning rate.

\begin{figure}
    \centering
    \includegraphics[width=\linewidth]{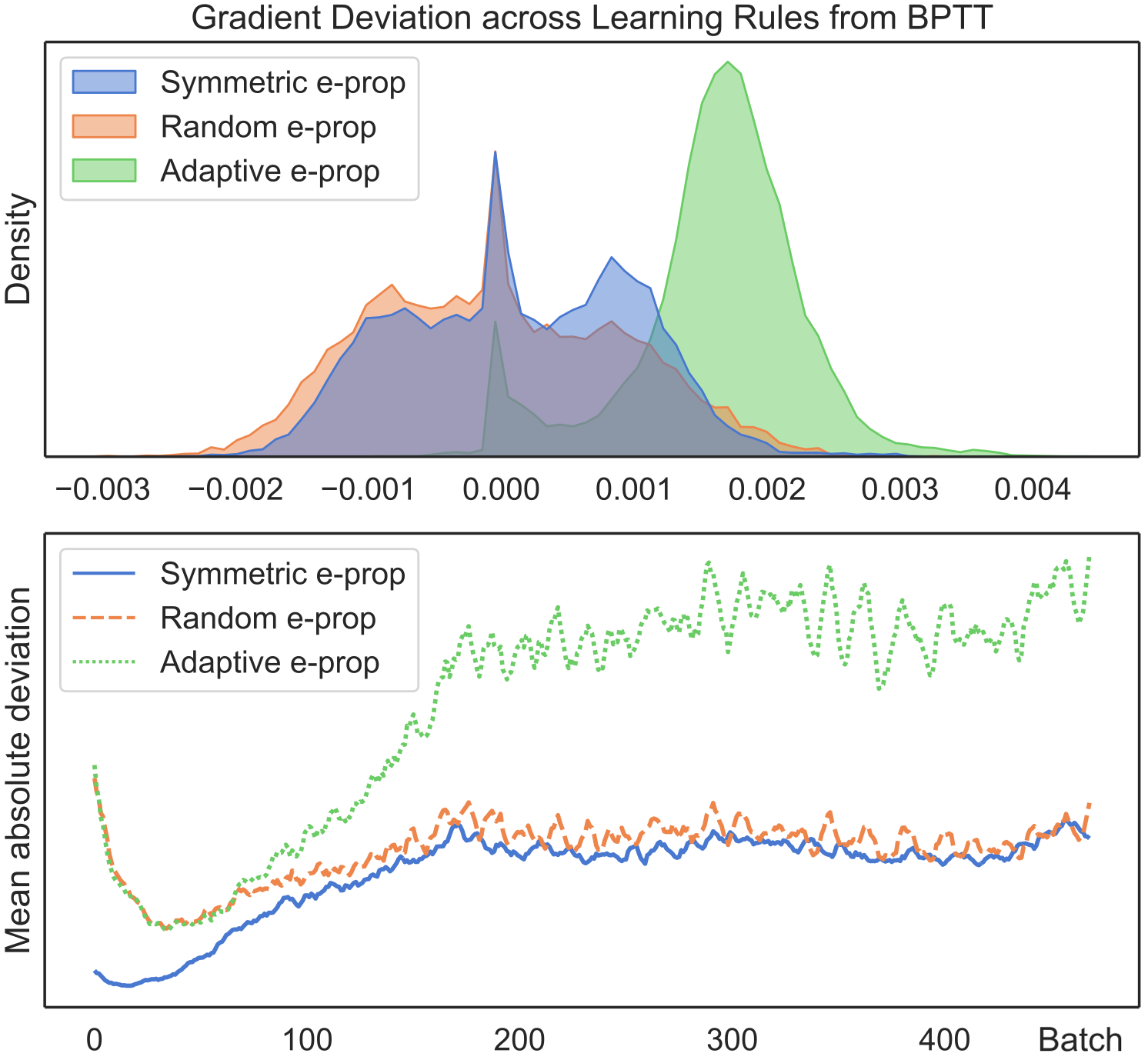}
    \caption{Gradient analysis tool.
      \textmd{This example illustrates the differences in approximate gradients among e-prop variants for training MNIST: (top) a snapshot of the distributions of gradient deviations, (bottom) how the gradient deviations change over time.}
    }
    \label{fig:grads}
\end{figure}

Gradient analysis offers insights into how learning rules behave relative to each other and backpropagation.
Fig. \ref{fig:grads} illustrates this with an example of training spiking MNIST models with three variants of e-prop.
While symmetric e-prop was the best at gradient approximation, the relationship between random and adaptive versions was somewhat unexpected.
The adaptive version produced gradients with larger deviation and bias, which could explain its weaker performance on the benchmark (not shown).

\section{Supporting utilities}
To further enable neuromorphic centric exploration, we integrate the SpikeCoding toolbox \cite{SpikeCoding2021} which enables simple encoding of continuous value sequences into spikes with nearly a dozen algorithms.
We present experimental results (Table \ref{tab:surg-ecg_table}) on two temporal data applications using three encoding schemes \cite{Petro2020}:
\begin{itemize}
\item \emph{Temporal contrast (TC)} encoding compares the absolute value of a signal with a threshold derived by the derivative and standard deviation of the full sequence multiplied by a tunable parameter.
\item \emph{Step-forward (SF)} encoding generates positive/negative spikes by comparing values in a sequence to a moving baseline plus a tunable threshold, which is initially the first value of the sequence and updated each spike.
\item \emph{Moving window (MW)} encoding uses a similar moving baseline and threshold to determine spiking but which is set to the mean of values in a tunable time window.
\end{itemize}

All models were trained with e-prop learning except for the Benchmark RNN model trained with BPTT.
While we note that there was often a sizable decrease in accuracy using these encodings, the sparsity of the input signal was significantly increased.
Spike encodings may enable the use and development of learning algorithms more suited to or dependent on event based input.

\begin{table}
\setlength{\tabcolsep}{7pt}
\centering
\caption{Testing two classification exemplars using temporal spike encoding schemes}
\label{tab:surg-ecg_table}
\begin{tabular}{lccccc}
\toprule
Encoding & None            & TC & SF & MW & Benchmark       \\
\midrule
Surgery$^{1}$  & 0.675           & 0.620   & 0.687   & 0.563   & \textbf{0.766} \\
ECG$^{2}$      & \textbf{0.813}  & 0.763   & 0.699   & 0.685   & 0.811    \\
\bottomrule
\end{tabular}
\begin{flushleft}
$^{1}$A surgery kinematic dataset measuring the positions and orientations of surgical instruments during labeled simulated exercises. Data available upon request.

$^{2}$A public ECG heartbeat categorization dataset \cite{kachuee2018ecg} subsampled for class balance.
\end{flushleft}
\end{table}

\section{Conclusions}
We presented the design of a coding library for researching learning algorithms.
Through three examples, we demonstrated its capability and ease of use in diverse scenarios.
Our reference implementations introduced a new state-of-the-art in local temporal credit assignment with SNNs, a sampling-based learning rule for estimating weight and prediction posteriors, as well as simulations for constrained training of analog neural networks on memristive hardware.
Additionally, we showed a cross-cutting example to support learning rule inspection with gradient comparison analysis.

Two directions emerge for future work.
First, we will extend learning rules to complex neuron models (e.g., dendritic computation, structured neurons) and network architecture.
Second, we will port learning algorithms to emerging hardware platforms.
Both processes will be facilitated by the abstraction of learning algorithms and the multi-backend support in the Neko library\footnote{https://github.com/cortical-team/neko}.

\pagebreak
\begin{acks}
We thank Sihong Wang and Shilei Dai for helpful discussions. 
This work is partially supported by Laboratory Directed Research and Development (LDRD) funding from Argonne National Laboratory, provided by the Director, Office of Science, of the U.S. Department of Energy under Contract No. DE-AC02-06CH11357.
\end{acks}

\bibliographystyle{ACM-Reference-Format}
\bibliography{main}


\begin{thebibliography}{45}


\ifx \showCODEN    \undefined \def \showCODEN     #1{\unskip}     \fi
\ifx \showDOI      \undefined \def \showDOI       #1{#1}\fi
\ifx \showISBNx    \undefined \def \showISBNx     #1{\unskip}     \fi
\ifx \showISBNxiii \undefined \def \showISBNxiii  #1{\unskip}     \fi
\ifx \showISSN     \undefined \def \showISSN      #1{\unskip}     \fi
\ifx \showLCCN     \undefined \def \showLCCN      #1{\unskip}     \fi
\ifx \shownote     \undefined \def \shownote      #1{#1}          \fi
\ifx \showarticletitle \undefined \def \showarticletitle #1{#1}   \fi
\ifx \showURL      \undefined \def \showURL       {\relax}        \fi
\providecommand\bibfield[2]{#2}
\providecommand\bibinfo[2]{#2}
\providecommand\natexlab[1]{#1}
\providecommand\showeprint[2][]{arXiv:#2}

\bibitem[\protect\citeauthoryear{Abadi, Barham, Chen, Chen, Davis, Dean, Devin,
  Ghemawat, Irving, Isard, et~al\mbox{.}}{Abadi et~al\mbox{.}}{2016}]%
        {abadi2016tensorflow}
\bibfield{author}{\bibinfo{person}{Mart{\'\i}n Abadi}, \bibinfo{person}{Paul
  Barham}, \bibinfo{person}{Jianmin Chen}, \bibinfo{person}{Zhifeng Chen},
  \bibinfo{person}{Andy Davis}, \bibinfo{person}{Jeffrey Dean},
  \bibinfo{person}{Matthieu Devin}, \bibinfo{person}{Sanjay Ghemawat},
  \bibinfo{person}{Geoffrey Irving}, \bibinfo{person}{Michael Isard},
  {et~al\mbox{.}}} \bibinfo{year}{2016}\natexlab{}.
\newblock \showarticletitle{Tensorflow: A system for large-scale machine
  learning}. In \bibinfo{booktitle}{\emph{12th $\{$USENIX$\}$ symposium on
  operating systems design and implementation ($\{$OSDI$\}$ 16)}}.
  \bibinfo{pages}{265--283}.
\newblock


\bibitem[\protect\citeauthoryear{Aitchison, Jegminat, Menendez, Pfister,
  Pouget, and Latham}{Aitchison et~al\mbox{.}}{2021}]%
        {aitchison2021synaptic}
\bibfield{author}{\bibinfo{person}{Laurence Aitchison}, \bibinfo{person}{Jannes
  Jegminat}, \bibinfo{person}{Jorge~Aurelio Menendez},
  \bibinfo{person}{Jean-Pascal Pfister}, \bibinfo{person}{Alexandre Pouget},
  {and} \bibinfo{person}{Peter~E Latham}.} \bibinfo{year}{2021}\natexlab{}.
\newblock \showarticletitle{Synaptic plasticity as Bayesian inference}.
\newblock \bibinfo{journal}{\emph{{Nature Neuroscience}}} \bibinfo{volume}{24},
  \bibinfo{number}{4} (\bibinfo{year}{2021}), \bibinfo{pages}{565--571}.
\newblock


\bibitem[\protect\citeauthoryear{Akrout, Wilson, Humphreys, Lillicrap, and
  Tweed}{Akrout et~al\mbox{.}}{2019}]%
        {akrout2019deep}
\bibfield{author}{\bibinfo{person}{Mohamed Akrout}, \bibinfo{person}{Collin
  Wilson}, \bibinfo{person}{Peter~C Humphreys}, \bibinfo{person}{Timothy
  Lillicrap}, {and} \bibinfo{person}{Douglas Tweed}.}
  \bibinfo{year}{2019}\natexlab{}.
\newblock \showarticletitle{Deep learning without weight transport}.
\newblock \bibinfo{journal}{\emph{arXiv preprint arXiv:1904.05391}}
  (\bibinfo{year}{2019}).
\newblock


\bibitem[\protect\citeauthoryear{Alibart, Gao, Hoskins, and Strukov}{Alibart
  et~al\mbox{.}}{2012}]%
        {Alibart_2012}
\bibfield{author}{\bibinfo{person}{Fabien Alibart}, \bibinfo{person}{Ligang
  Gao}, \bibinfo{person}{Brian~D Hoskins}, {and} \bibinfo{person}{Dmitri~B
  Strukov}.} \bibinfo{year}{2012}\natexlab{}.
\newblock \showarticletitle{High precision tuning of state for memristive
  devices by adaptable variation-tolerant algorithm}.
\newblock \bibinfo{journal}{\emph{Nanotechnology}} \bibinfo{volume}{23},
  \bibinfo{number}{7} (\bibinfo{date}{Jan} \bibinfo{year}{2012}),
  \bibinfo{pages}{075201}.
\newblock
\showISSN{1361-6528}
\urldef\tempurl%
\url{https://doi.org/10.1088/0957-4484/23/7/075201}
\showDOI{\tempurl}


\bibitem[\protect\citeauthoryear{Andrieu and Thoms}{Andrieu and Thoms}{2008}]%
        {Andrieu2008Adaptive}
\bibfield{author}{\bibinfo{person}{Christophe Andrieu} {and}
  \bibinfo{person}{Johannes Thoms}.} \bibinfo{year}{2008}\natexlab{}.
\newblock \showarticletitle{A tutorial on adaptive MCMC}.
\newblock \bibinfo{journal}{\emph{Statistics and Computing}}
  \bibinfo{volume}{18}, \bibinfo{number}{4} (\bibinfo{date}{01 Dec}
  \bibinfo{year}{2008}), \bibinfo{pages}{343--373}.
\newblock
\showISSN{1573-1375}
\urldef\tempurl%
\url{https://doi.org/10.1007/s11222-008-9110-y}
\showDOI{\tempurl}


\bibitem[\protect\citeauthoryear{Bekolay, Bergstra, Hunsberger, DeWolf,
  Stewart, Rasmussen, Choo, Voelker, and Eliasmith}{Bekolay
  et~al\mbox{.}}{2014}]%
        {bekolay2014nengo}
\bibfield{author}{\bibinfo{person}{Trevor Bekolay}, \bibinfo{person}{James
  Bergstra}, \bibinfo{person}{Eric Hunsberger}, \bibinfo{person}{Travis
  DeWolf}, \bibinfo{person}{Terrence~C Stewart}, \bibinfo{person}{Daniel
  Rasmussen}, \bibinfo{person}{Xuan Choo}, \bibinfo{person}{Aaron Voelker},
  {and} \bibinfo{person}{Chris Eliasmith}.} \bibinfo{year}{2014}\natexlab{}.
\newblock \showarticletitle{Nengo: a Python tool for building large-scale
  functional brain models}.
\newblock \bibinfo{journal}{\emph{{Frontiers in Neuroinformatics}}}
  \bibinfo{volume}{7} (\bibinfo{year}{2014}), \bibinfo{pages}{48}.
\newblock


\bibitem[\protect\citeauthoryear{Bellec, Scherr, Subramoney, Hajek, Salaj,
  Legenstein, and Maass}{Bellec et~al\mbox{.}}{2020}]%
        {bellec2020solution}
\bibfield{author}{\bibinfo{person}{Guillaume Bellec}, \bibinfo{person}{Franz
  Scherr}, \bibinfo{person}{Anand Subramoney}, \bibinfo{person}{Elias Hajek},
  \bibinfo{person}{Darjan Salaj}, \bibinfo{person}{Robert Legenstein}, {and}
  \bibinfo{person}{Wolfgang Maass}.} \bibinfo{year}{2020}\natexlab{}.
\newblock \showarticletitle{A solution to the learning dilemma for recurrent
  networks of spiking neurons}.
\newblock \bibinfo{journal}{\emph{{Nature Communications}}}
  \bibinfo{volume}{11}, \bibinfo{number}{1} (\bibinfo{year}{2020}),
  \bibinfo{pages}{1--15}.
\newblock


\bibitem[\protect\citeauthoryear{Carnevale and Hines}{Carnevale and
  Hines}{2006}]%
        {carnevale2006neuron}
\bibfield{author}{\bibinfo{person}{Nicholas~T Carnevale} {and}
  \bibinfo{person}{Michael~L Hines}.} \bibinfo{year}{2006}\natexlab{}.
\newblock \bibinfo{booktitle}{\emph{{The NEURON book}}}.
\newblock \bibinfo{publisher}{Cambridge University Press}.
\newblock


\bibitem[\protect\citeauthoryear{Chollet et~al\mbox{.}}{Chollet
  et~al\mbox{.}}{2015}]%
        {chollet2015keras}
\bibfield{author}{\bibinfo{person}{Fran\c{c}ois Chollet} {et~al\mbox{.}}}
  \bibinfo{year}{2015}\natexlab{}.
\newblock \bibinfo{title}{Keras}.
\newblock \bibinfo{howpublished}{\url{https://keras.io}}.
\newblock


\bibitem[\protect\citeauthoryear{Dabney, Kurth-Nelson, Uchida, Starkweather,
  Hassabis, Munos, and Botvinick}{Dabney et~al\mbox{.}}{2020}]%
        {dabney2020distributional}
\bibfield{author}{\bibinfo{person}{Will Dabney}, \bibinfo{person}{Zeb
  Kurth-Nelson}, \bibinfo{person}{Naoshige Uchida}, \bibinfo{person}{Clara~Kwon
  Starkweather}, \bibinfo{person}{Demis Hassabis}, \bibinfo{person}{R{\'e}mi
  Munos}, {and} \bibinfo{person}{Matthew Botvinick}.}
  \bibinfo{year}{2020}\natexlab{}.
\newblock \showarticletitle{A distributional code for value in dopamine-based
  reinforcement learning}.
\newblock \bibinfo{journal}{\emph{Nature}} \bibinfo{volume}{577},
  \bibinfo{number}{7792} (\bibinfo{year}{2020}), \bibinfo{pages}{671--675}.
\newblock


\bibitem[\protect\citeauthoryear{Davison, Br{\"u}derle, Eppler, Kremkow,
  Muller, Pecevski, Perrinet, and Yger}{Davison et~al\mbox{.}}{2009}]%
        {davison2009pynn}
\bibfield{author}{\bibinfo{person}{Andrew~P Davison}, \bibinfo{person}{Daniel
  Br{\"u}derle}, \bibinfo{person}{Jochen~M Eppler}, \bibinfo{person}{Jens
  Kremkow}, \bibinfo{person}{Eilif Muller}, \bibinfo{person}{Dejan Pecevski},
  \bibinfo{person}{Laurent Perrinet}, {and} \bibinfo{person}{Pierre Yger}.}
  \bibinfo{year}{2009}\natexlab{}.
\newblock \showarticletitle{PyNN: a common interface for neuronal network
  simulators}.
\newblock \bibinfo{journal}{\emph{{Frontiers in Neuroinformatics}}}
  \bibinfo{volume}{2} (\bibinfo{year}{2009}), \bibinfo{pages}{11}.
\newblock


\bibitem[\protect\citeauthoryear{Dupeyroux}{Dupeyroux}{2021}]%
        {SpikeCoding2021}
\bibfield{author}{\bibinfo{person}{Julien Dupeyroux}.}
  \bibinfo{year}{2021}\natexlab{}.
\newblock \bibinfo{title}{A toolbox for neuromorphic sensing in robotics}.
\newblock
\newblock
\showeprint[arxiv]{2103.02751}~[cs.RO]


\bibitem[\protect\citeauthoryear{Fuller, Keene, Melianas, Wang, Agarwal, Li,
  Tuchman, James, Marinella, Yang, et~al\mbox{.}}{Fuller et~al\mbox{.}}{2019}]%
        {fuller2019parallel}
\bibfield{author}{\bibinfo{person}{Elliot~J Fuller}, \bibinfo{person}{Scott~T
  Keene}, \bibinfo{person}{Armantas Melianas}, \bibinfo{person}{Zhongrui Wang},
  \bibinfo{person}{Sapan Agarwal}, \bibinfo{person}{Yiyang Li},
  \bibinfo{person}{Yaakov Tuchman}, \bibinfo{person}{Conrad~D James},
  \bibinfo{person}{Matthew~J Marinella}, \bibinfo{person}{J~Joshua Yang},
  {et~al\mbox{.}}} \bibinfo{year}{2019}\natexlab{}.
\newblock \showarticletitle{Parallel programming of an ionic floating-gate
  memory array for scalable neuromorphic computing}.
\newblock \bibinfo{journal}{\emph{Science}} \bibinfo{volume}{364},
  \bibinfo{number}{6440} (\bibinfo{year}{2019}), \bibinfo{pages}{570--574}.
\newblock


\bibitem[\protect\citeauthoryear{Gal and Ghahramani}{Gal and
  Ghahramani}{2016}]%
        {gal2016}
\bibfield{author}{\bibinfo{person}{Yarin Gal} {and} \bibinfo{person}{Zoubin
  Ghahramani}.} \bibinfo{year}{2016}\natexlab{}.
\newblock \showarticletitle{Dropout as a Bayesian Approximation: Representing
  Model Uncertainty in Deep Learning}. In \bibinfo{booktitle}{\emph{Proceedings
  of The 33rd International Conference on Machine Learning}}
  \emph{(\bibinfo{series}{Proceedings of Machine Learning Research},
  Vol.~\bibinfo{volume}{48})}, \bibfield{editor}{\bibinfo{person}{Maria~Florina
  Balcan} {and} \bibinfo{person}{Kilian~Q. Weinberger}} (Eds.).
  \bibinfo{publisher}{PMLR}, \bibinfo{address}{New York, New York, USA},
  \bibinfo{pages}{1050--1059}.
\newblock
\urldef\tempurl%
\url{http://proceedings.mlr.press/v48/gal16.html}
\showURL{%
\tempurl}


\bibitem[\protect\citeauthoryear{Garofolo, Lamel, Fisher, Fiscus, Pallett,
  Dahlgren, and Zue}{Garofolo et~al\mbox{.}}{1992}]%
        {garofolo1992timit}
\bibfield{author}{\bibinfo{person}{J. Garofolo}, \bibinfo{person}{Lori Lamel},
  \bibinfo{person}{W. Fisher}, \bibinfo{person}{Jonathan Fiscus},
  \bibinfo{person}{D. Pallett}, \bibinfo{person}{N. Dahlgren}, {and}
  \bibinfo{person}{V. Zue}.} \bibinfo{year}{1992}\natexlab{}.
\newblock \showarticletitle{TIMIT Acoustic-phonetic Continuous Speech Corpus}.
\newblock \bibinfo{journal}{\emph{Linguistic Data Consortium}}
  (\bibinfo{date}{11} \bibinfo{year}{1992}).
\newblock


\bibitem[\protect\citeauthoryear{Gewaltig and Diesmann}{Gewaltig and
  Diesmann}{2007}]%
        {gewaltig2007nest}
\bibfield{author}{\bibinfo{person}{Marc-Oliver Gewaltig} {and}
  \bibinfo{person}{Markus Diesmann}.} \bibinfo{year}{2007}\natexlab{}.
\newblock \showarticletitle{Nest (neural simulation tool)}.
\newblock \bibinfo{journal}{\emph{Scholarpedia}} \bibinfo{volume}{2},
  \bibinfo{number}{4} (\bibinfo{year}{2007}), \bibinfo{pages}{1430}.
\newblock


\bibitem[\protect\citeauthoryear{Graves}{Graves}{2011}]%
        {Graves2011}
\bibfield{author}{\bibinfo{person}{Alex Graves}.}
  \bibinfo{year}{2011}\natexlab{}.
\newblock \showarticletitle{Practical Variational Inference for Neural
  Networks}. In \bibinfo{booktitle}{\emph{Proceedings of the 24th International
  Conference on Neural Information Processing Systems}} (Granada, Spain)
  \emph{(\bibinfo{series}{NIPS'11})}. \bibinfo{publisher}{Curran Associates
  Inc.}, \bibinfo{address}{Red Hook, NY, USA}, \bibinfo{pages}{2348–2356}.
\newblock
\showISBNx{9781618395993}


\bibitem[\protect\citeauthoryear{Greydanus}{Greydanus}{2020}]%
        {greydanus2020scaling}
\bibfield{author}{\bibinfo{person}{Sam Greydanus}.}
  \bibinfo{year}{2020}\natexlab{}.
\newblock \bibinfo{title}{Scaling down Deep Learning}.
\newblock
\newblock
\showeprint[arxiv]{2011.14439}~[cs.LG]


\bibitem[\protect\citeauthoryear{Hoff}{Hoff}{2009}]%
        {Hoff2009Bayes}
\bibfield{author}{\bibinfo{person}{Peter~D. Hoff}.}
  \bibinfo{year}{2009}\natexlab{}.
\newblock \bibinfo{booktitle}{\emph{A First Course in Bayesian Statistical
  Methods} (\bibinfo{edition}{1st} ed.)}.
\newblock \bibinfo{publisher}{Springer Publishing Company, Incorporated}.
\newblock
\showISBNx{0387922997}


\bibitem[\protect\citeauthoryear{Hu, Li, Chen, Wu, Rose, and Linderman}{Hu
  et~al\mbox{.}}{2014}]%
        {hu2014memristor}
\bibfield{author}{\bibinfo{person}{Miao Hu}, \bibinfo{person}{Hai Li},
  \bibinfo{person}{Yiran Chen}, \bibinfo{person}{Qing Wu},
  \bibinfo{person}{Garrett~S Rose}, {and} \bibinfo{person}{Richard~W
  Linderman}.} \bibinfo{year}{2014}\natexlab{}.
\newblock \showarticletitle{Memristor crossbar-based neuromorphic computing
  system: A case study}.
\newblock \bibinfo{journal}{\emph{{IEEE Transactions on Neural Networks and
  Learning Systems}}} \bibinfo{volume}{25}, \bibinfo{number}{10}
  (\bibinfo{year}{2014}), \bibinfo{pages}{1864--1878}.
\newblock


\bibitem[\protect\citeauthoryear{Kachuee, Fazeli, and Sarrafzadeh}{Kachuee
  et~al\mbox{.}}{2018}]%
        {kachuee2018ecg}
\bibfield{author}{\bibinfo{person}{Mohammad Kachuee}, \bibinfo{person}{Shayan
  Fazeli}, {and} \bibinfo{person}{Majid Sarrafzadeh}.}
  \bibinfo{year}{2018}\natexlab{}.
\newblock \showarticletitle{Ecg heartbeat classification: A deep transferable
  representation}. In \bibinfo{booktitle}{\emph{2018 IEEE International
  Conference on Healthcare Informatics (ICHI)}}. IEEE,
  \bibinfo{pages}{443--444}.
\newblock


\bibitem[\protect\citeauthoryear{Knill and Pouget}{Knill and Pouget}{2004}]%
        {Knill2004}
\bibfield{author}{\bibinfo{person}{David~C. Knill} {and}
  \bibinfo{person}{Alexandre Pouget}.} \bibinfo{year}{2004}\natexlab{}.
\newblock \showarticletitle{The Bayesian brain: the role of uncertainty in
  neural coding and computation}.
\newblock \bibinfo{journal}{\emph{Trends in Neurosciences}}
  \bibinfo{volume}{27}, \bibinfo{number}{12} (\bibinfo{date}{01 Dec}
  \bibinfo{year}{2004}), \bibinfo{pages}{712--719}.
\newblock
\showISSN{0166-2236}
\urldef\tempurl%
\url{https://doi.org/10.1016/j.tins.2004.10.007}
\showDOI{\tempurl}


\bibitem[\protect\citeauthoryear{LeCun}{LeCun}{1998}]%
        {lecun1998mnist}
\bibfield{author}{\bibinfo{person}{Yann LeCun}.}
  \bibinfo{year}{1998}\natexlab{}.
\newblock \showarticletitle{The MNIST database of handwritten digits}.
\newblock \bibinfo{journal}{\emph{http://yann. lecun. com/exdb/mnist/}}
  (\bibinfo{year}{1998}).
\newblock


\bibitem[\protect\citeauthoryear{Lee, Delbruck, and Pfeiffer}{Lee
  et~al\mbox{.}}{2016}]%
        {lee2016training}
\bibfield{author}{\bibinfo{person}{Jun~Haeng Lee}, \bibinfo{person}{Tobi
  Delbruck}, {and} \bibinfo{person}{Michael Pfeiffer}.}
  \bibinfo{year}{2016}\natexlab{}.
\newblock \showarticletitle{Training deep spiking neural networks using
  backpropagation}.
\newblock \bibinfo{journal}{\emph{{Frontiers in Neuroscience}}}
  \bibinfo{volume}{10} (\bibinfo{year}{2016}), \bibinfo{pages}{508}.
\newblock


\bibitem[\protect\citeauthoryear{Li, Belkin, Li, Yan, Hu, Ge, Jiang,
  Montgomery, Lin, Wang, et~al\mbox{.}}{Li et~al\mbox{.}}{2018}]%
        {li2018efficient}
\bibfield{author}{\bibinfo{person}{Can Li}, \bibinfo{person}{Daniel Belkin},
  \bibinfo{person}{Yunning Li}, \bibinfo{person}{Peng Yan},
  \bibinfo{person}{Miao Hu}, \bibinfo{person}{Ning Ge}, \bibinfo{person}{Hao
  Jiang}, \bibinfo{person}{Eric Montgomery}, \bibinfo{person}{Peng Lin},
  \bibinfo{person}{Zhongrui Wang}, {et~al\mbox{.}}}
  \bibinfo{year}{2018}\natexlab{}.
\newblock \showarticletitle{Efficient and self-adaptive in-situ learning in
  multilayer memristor neural networks}.
\newblock \bibinfo{journal}{\emph{Nature communications}} \bibinfo{volume}{9},
  \bibinfo{number}{1} (\bibinfo{year}{2018}), \bibinfo{pages}{1--8}.
\newblock


\bibitem[\protect\citeauthoryear{Lillicrap, Cownden, Tweed, and
  Akerman}{Lillicrap et~al\mbox{.}}{2016}]%
        {lillicrap2016random}
\bibfield{author}{\bibinfo{person}{Timothy~P Lillicrap},
  \bibinfo{person}{Daniel Cownden}, \bibinfo{person}{Douglas~B Tweed}, {and}
  \bibinfo{person}{Colin~J Akerman}.} \bibinfo{year}{2016}\natexlab{}.
\newblock \showarticletitle{Random synaptic feedback weights support error
  backpropagation for deep learning}.
\newblock \bibinfo{journal}{\emph{Nature communications}} \bibinfo{volume}{7},
  \bibinfo{number}{1} (\bibinfo{year}{2016}), \bibinfo{pages}{1--10}.
\newblock


\bibitem[\protect\citeauthoryear{Lillicrap, Santoro, Marris, Akerman, and
  Hinton}{Lillicrap et~al\mbox{.}}{2020}]%
        {lillicrap2020backpropagation}
\bibfield{author}{\bibinfo{person}{Timothy~P Lillicrap}, \bibinfo{person}{Adam
  Santoro}, \bibinfo{person}{Luke Marris}, \bibinfo{person}{Colin~J Akerman},
  {and} \bibinfo{person}{Geoffrey Hinton}.} \bibinfo{year}{2020}\natexlab{}.
\newblock \showarticletitle{Backpropagation and the brain}.
\newblock \bibinfo{journal}{\emph{{Nature Reviews Neuroscience}}}
  \bibinfo{volume}{21}, \bibinfo{number}{6} (\bibinfo{year}{2020}),
  \bibinfo{pages}{335--346}.
\newblock


\bibitem[\protect\citeauthoryear{Lin, Wild, Chinya, Cao, Davies, Lavery, and
  Wang}{Lin et~al\mbox{.}}{2018}]%
        {lin2018programming}
\bibfield{author}{\bibinfo{person}{Chit-Kwan Lin}, \bibinfo{person}{Andreas
  Wild}, \bibinfo{person}{Gautham~N Chinya}, \bibinfo{person}{Yongqiang Cao},
  \bibinfo{person}{Mike Davies}, \bibinfo{person}{Daniel~M Lavery}, {and}
  \bibinfo{person}{Hong Wang}.} \bibinfo{year}{2018}\natexlab{}.
\newblock \showarticletitle{Programming spiking neural networks on Intel’s
  Loihi}.
\newblock \bibinfo{journal}{\emph{Computer}} \bibinfo{volume}{51},
  \bibinfo{number}{3} (\bibinfo{year}{2018}), \bibinfo{pages}{52--61}.
\newblock


\bibitem[\protect\citeauthoryear{Marschall, Cho, and Savin}{Marschall
  et~al\mbox{.}}{2020}]%
        {marschall2020unified}
\bibfield{author}{\bibinfo{person}{Owen Marschall}, \bibinfo{person}{Kyunghyun
  Cho}, {and} \bibinfo{person}{Cristina Savin}.}
  \bibinfo{year}{2020}\natexlab{}.
\newblock \showarticletitle{A unified framework of online learning algorithms
  for training recurrent neural networks}.
\newblock \bibinfo{journal}{\emph{Journal of Machine Learning Research}}
  \bibinfo{volume}{21}, \bibinfo{number}{135} (\bibinfo{year}{2020}),
  \bibinfo{pages}{1--34}.
\newblock


\bibitem[\protect\citeauthoryear{Neal}{Neal}{2011}]%
        {Neil2011HMC}
\bibfield{author}{\bibinfo{person}{Radford~M. Neal}.}
  \bibinfo{year}{2011}\natexlab{}.
\newblock \bibinfo{booktitle}{\emph{MCMC Using Hamiltonian Dynamics}}.
\newblock \bibinfo{publisher}{CRC Press}.
\newblock
\urldef\tempurl%
\url{https://doi.org/10.1201/b10905-7}
\showDOI{\tempurl}


\bibitem[\protect\citeauthoryear{Neftci, Mostafa, and Zenke}{Neftci
  et~al\mbox{.}}{2019}]%
        {neftci2019surrogate}
\bibfield{author}{\bibinfo{person}{Emre~O Neftci}, \bibinfo{person}{Hesham
  Mostafa}, {and} \bibinfo{person}{Friedemann Zenke}.}
  \bibinfo{year}{2019}\natexlab{}.
\newblock \showarticletitle{Surrogate gradient learning in spiking neural
  networks: Bringing the power of gradient-based optimization to spiking neural
  networks}.
\newblock \bibinfo{journal}{\emph{IEEE Signal Processing Magazine}}
  \bibinfo{volume}{36}, \bibinfo{number}{6} (\bibinfo{year}{2019}),
  \bibinfo{pages}{51--63}.
\newblock


\bibitem[\protect\citeauthoryear{Paszke, Gross, Massa, Lerer, Bradbury, Chanan,
  Killeen, Lin, Gimelshein, Antiga, et~al\mbox{.}}{Paszke
  et~al\mbox{.}}{[n.d.]}]%
        {paszke2019pytorch}
\bibfield{author}{\bibinfo{person}{Adam Paszke}, \bibinfo{person}{Sam Gross},
  \bibinfo{person}{Francisco Massa}, \bibinfo{person}{Adam Lerer},
  \bibinfo{person}{James Bradbury}, \bibinfo{person}{Gregory Chanan},
  \bibinfo{person}{Trevor Killeen}, \bibinfo{person}{Zeming Lin},
  \bibinfo{person}{Natalia Gimelshein}, \bibinfo{person}{Luca Antiga},
  {et~al\mbox{.}}} \bibinfo{year}{[n.d.]}\natexlab{}.
\newblock \showarticletitle{PyTorch: An imperative style, high-performance deep
  learning library}.
\newblock \bibinfo{journal}{\emph{arXiv preprint arXiv:1912.01703}}
  (\bibinfo{year}{[n.\,d.]}).
\newblock


\bibitem[\protect\citeauthoryear{Petro, Kasabov, and Kiss}{Petro
  et~al\mbox{.}}{2020}]%
        {Petro2020}
\bibfield{author}{\bibinfo{person}{Balint Petro}, \bibinfo{person}{Nikola
  Kasabov}, {and} \bibinfo{person}{Rita~M. Kiss}.}
  \bibinfo{year}{2020}\natexlab{}.
\newblock \showarticletitle{Selection and Optimization of Temporal Spike
  Encoding Methods for Spiking Neural Networks}.
\newblock \bibinfo{journal}{\emph{{IEEE} Transactions on Neural Networks and
  Learning Systems}} \bibinfo{volume}{31}, \bibinfo{number}{2}
  (\bibinfo{date}{Feb.} \bibinfo{year}{2020}), \bibinfo{pages}{358--370}.
\newblock
\urldef\tempurl%
\url{https://doi.org/10.1109/tnnls.2019.2906158}
\showDOI{\tempurl}


\bibitem[\protect\citeauthoryear{Rossky, Doll, and Friedman}{Rossky
  et~al\mbox{.}}{1978}]%
        {Rossky1978}
\bibfield{author}{\bibinfo{person}{P.~J. Rossky}, \bibinfo{person}{J.~D. Doll},
  {and} \bibinfo{person}{H.~L. Friedman}.} \bibinfo{year}{1978}\natexlab{}.
\newblock \showarticletitle{Brownian dynamics as smart Monte Carlo simulation}.
\newblock \bibinfo{journal}{\emph{The Journal of Chemical Physics}}
  \bibinfo{volume}{69}, \bibinfo{number}{10} (\bibinfo{year}{1978}),
  \bibinfo{pages}{4628--4633}.
\newblock
\urldef\tempurl%
\url{https://doi.org/10.1063/1.436415}
\showDOI{\tempurl}
\showeprint{https://doi.org/10.1063/1.436415}


\bibitem[\protect\citeauthoryear{Rueckauer, Bybee, Goettsche, Singh, Mishra,
  and Wild}{Rueckauer et~al\mbox{.}}{2021}]%
        {rueckauer2021nxtf}
\bibfield{author}{\bibinfo{person}{Bodo Rueckauer}, \bibinfo{person}{Connor
  Bybee}, \bibinfo{person}{Ralf Goettsche}, \bibinfo{person}{Yashwardhan
  Singh}, \bibinfo{person}{Joyesh Mishra}, {and} \bibinfo{person}{Andreas
  Wild}.} \bibinfo{year}{2021}\natexlab{}.
\newblock \showarticletitle{NxTF: An API and Compiler for Deep Spiking Neural
  Networks on Intel Loihi}.
\newblock \bibinfo{journal}{\emph{arXiv preprint arXiv:2101.04261}}
  (\bibinfo{year}{2021}).
\newblock


\bibitem[\protect\citeauthoryear{Rueckauer and Liu}{Rueckauer and Liu}{2018}]%
        {rueckauer2018conversion}
\bibfield{author}{\bibinfo{person}{Bodo Rueckauer} {and}
  \bibinfo{person}{Shih-Chii Liu}.} \bibinfo{year}{2018}\natexlab{}.
\newblock \showarticletitle{Conversion of analog to spiking neural networks
  using sparse temporal coding}. In \bibinfo{booktitle}{\emph{2018 IEEE
  International Symposium on Circuits and Systems (ISCAS)}}. IEEE,
  \bibinfo{pages}{1--5}.
\newblock


\bibitem[\protect\citeauthoryear{Rueckauer, Lungu, Hu, Pfeiffer, and
  Liu}{Rueckauer et~al\mbox{.}}{2017}]%
        {rueckauer2017conversion}
\bibfield{author}{\bibinfo{person}{Bodo Rueckauer},
  \bibinfo{person}{Iulia-Alexandra Lungu}, \bibinfo{person}{Yuhuang Hu},
  \bibinfo{person}{Michael Pfeiffer}, {and} \bibinfo{person}{Shih-Chii Liu}.}
  \bibinfo{year}{2017}\natexlab{}.
\newblock \showarticletitle{Conversion of continuous-valued deep networks to
  efficient event-driven networks for image classification}.
\newblock \bibinfo{journal}{\emph{{Frontiers in Neuroscience}}}
  \bibinfo{volume}{11} (\bibinfo{year}{2017}), \bibinfo{pages}{682}.
\newblock


\bibitem[\protect\citeauthoryear{Sacramento, Costa, Bengio, and
  Senn}{Sacramento et~al\mbox{.}}{2018}]%
        {sacramento2018dendritic}
\bibfield{author}{\bibinfo{person}{Jo{\~a}o Sacramento},
  \bibinfo{person}{Rui~Ponte Costa}, \bibinfo{person}{Yoshua Bengio}, {and}
  \bibinfo{person}{Walter Senn}.} \bibinfo{year}{2018}\natexlab{}.
\newblock \showarticletitle{Dendritic cortical microcircuits approximate the
  backpropagation algorithm}.
\newblock \bibinfo{journal}{\emph{arXiv preprint arXiv:1810.11393}}
  (\bibinfo{year}{2018}).
\newblock


\bibitem[\protect\citeauthoryear{Sawada, Akopyan, Cassidy, Taba, Debole, Datta,
  Alvarez-Icaza, Amir, Arthur, Andreopoulos, et~al\mbox{.}}{Sawada
  et~al\mbox{.}}{2016}]%
        {sawada2016truenorth}
\bibfield{author}{\bibinfo{person}{Jun Sawada}, \bibinfo{person}{Filipp
  Akopyan}, \bibinfo{person}{Andrew~S Cassidy}, \bibinfo{person}{Brian Taba},
  \bibinfo{person}{Michael~V Debole}, \bibinfo{person}{Pallab Datta},
  \bibinfo{person}{Rodrigo Alvarez-Icaza}, \bibinfo{person}{Arnon Amir},
  \bibinfo{person}{John~V Arthur}, \bibinfo{person}{Alexander Andreopoulos},
  {et~al\mbox{.}}} \bibinfo{year}{2016}\natexlab{}.
\newblock \showarticletitle{Truenorth ecosystem for brain-inspired computing:
  scalable systems, software, and applications}. In
  \bibinfo{booktitle}{\emph{SC'16: Proceedings of the International Conference
  for High Performance Computing, Networking, Storage and Analysis}}. IEEE,
  \bibinfo{pages}{130--141}.
\newblock


\bibitem[\protect\citeauthoryear{Sornborger, Tao, Snyder, and
  Zlotnik}{Sornborger et~al\mbox{.}}{2019}]%
        {sornborger2019pulse}
\bibfield{author}{\bibinfo{person}{Andrew Sornborger}, \bibinfo{person}{Louis
  Tao}, \bibinfo{person}{Jordan Snyder}, {and} \bibinfo{person}{Anatoly
  Zlotnik}.} \bibinfo{year}{2019}\natexlab{}.
\newblock \showarticletitle{A pulse-gated, neural implementation of the
  backpropagation algorithm}. In \bibinfo{booktitle}{\emph{Proceedings of the
  7th Annual Neuro-inspired Computational Elements Workshop}}.
  \bibinfo{pages}{1--9}.
\newblock


\bibitem[\protect\citeauthoryear{Stimberg, Brette, and Goodman}{Stimberg
  et~al\mbox{.}}{2019}]%
        {stimberg2019brian}
\bibfield{author}{\bibinfo{person}{Marcel Stimberg}, \bibinfo{person}{Romain
  Brette}, {and} \bibinfo{person}{Dan~FM Goodman}.}
  \bibinfo{year}{2019}\natexlab{}.
\newblock \showarticletitle{Brian 2, an intuitive and efficient neural
  simulator}.
\newblock \bibinfo{journal}{\emph{{eLife}}}  \bibinfo{volume}{8}
  (\bibinfo{year}{2019}), \bibinfo{pages}{e47314}.
\newblock


\bibitem[\protect\citeauthoryear{Thomas}{Thomas}{2013}]%
        {thomas2013memristor}
\bibfield{author}{\bibinfo{person}{Andy Thomas}.}
  \bibinfo{year}{2013}\natexlab{}.
\newblock \showarticletitle{Memristor-based neural networks}.
\newblock \bibinfo{journal}{\emph{{Journal of Physics D: Applied Physics}}}
  \bibinfo{volume}{46}, \bibinfo{number}{9} (\bibinfo{year}{2013}),
  \bibinfo{pages}{093001}.
\newblock


\bibitem[\protect\citeauthoryear{Yao, Wu, Gao, Tang, Zhang, Zhang, Yang, and
  Qian}{Yao et~al\mbox{.}}{2020}]%
        {yao2020fully}
\bibfield{author}{\bibinfo{person}{Peng Yao}, \bibinfo{person}{Huaqiang Wu},
  \bibinfo{person}{Bin Gao}, \bibinfo{person}{Jianshi Tang},
  \bibinfo{person}{Qingtian Zhang}, \bibinfo{person}{Wenqiang Zhang},
  \bibinfo{person}{J~Joshua Yang}, {and} \bibinfo{person}{He Qian}.}
  \bibinfo{year}{2020}\natexlab{}.
\newblock \showarticletitle{Fully hardware-implemented memristor convolutional
  neural network}.
\newblock \bibinfo{journal}{\emph{Nature}} \bibinfo{volume}{577},
  \bibinfo{number}{7792} (\bibinfo{year}{2020}), \bibinfo{pages}{641--646}.
\newblock


\bibitem[\protect\citeauthoryear{Zamanidoost, Bayat, Strukov, and
  Kataeva}{Zamanidoost et~al\mbox{.}}{2015}]%
        {7139171}
\bibfield{author}{\bibinfo{person}{Elham Zamanidoost},
  \bibinfo{person}{Farnood~M. Bayat}, \bibinfo{person}{Dmitri Strukov}, {and}
  \bibinfo{person}{Irina Kataeva}.} \bibinfo{year}{2015}\natexlab{}.
\newblock \showarticletitle{Manhattan rule training for memristive crossbar
  circuit pattern classifiers}. In \bibinfo{booktitle}{\emph{2015 IEEE 9th
  International Symposium on Intelligent Signal Processing (WISP)
  Proceedings}}. \bibinfo{pages}{1--6}.
\newblock
\urldef\tempurl%
\url{https://doi.org/10.1109/WISP.2015.7139171}
\showDOI{\tempurl}


\bibitem[\protect\citeauthoryear{Zenke and Ganguli}{Zenke and Ganguli}{2018}]%
        {zenke2018superspike}
\bibfield{author}{\bibinfo{person}{Friedemann Zenke} {and}
  \bibinfo{person}{Surya Ganguli}.} \bibinfo{year}{2018}\natexlab{}.
\newblock \showarticletitle{Superspike: Supervised learning in multilayer
  spiking neural networks}.
\newblock \bibinfo{journal}{\emph{Neural computation}} \bibinfo{volume}{30},
  \bibinfo{number}{6} (\bibinfo{year}{2018}), \bibinfo{pages}{1514--1541}.
\newblock


\end{thebibliography}





\end{document}